\documentclass[10pt,twocolumn,letterpaper]{article}

\usepackage[accsupp]{axessibility} % Improves PDF readability for those with disabilities.
\usepackage{cvpr}              % To produce the CAMERA-READY version
\usepackage{multirow}
\usepackage{arydshln} 
\usepackage{amsthm}
\usepackage{adjustbox} % For adjusting the figure's dimensions
\usepackage[dvipsnames]{xcolor}

\definecolor{cvprblue}{rgb}{0.21,0.49,0.74}
\usepackage[pagebackref,breaklinks,colorlinks,citecolor=cvprblue]{hyperref}
\usepackage{float}
\def\paperID{12} % *** Enter the Paper ID here
\def\confName{CVPR}
\def\confYear{2024}

\title{Interpreting COVID Lateral Flow Tests' Results with Foundation Models}

\author{Stuti Pandey\textsuperscript{1}, Josh Myers-Dean\textsuperscript{1}, Jarek Reynolds\textsuperscript{1}, Danna Gurari\textsuperscript{1,2}\\
{\tt\small \textsuperscript{1}University of Colorado Boulder, \textsuperscript{2}The University of Texas at Austin }\\
}

\begin{document}
\maketitle

%%%%%%%%% ABSTRACT
\begin{abstract}
Lateral flow tests (LFTs) enable rapid, low-cost testing for health conditions including Covid, pregnancy, HIV, and malaria.  Automated readers of LFT results can yield many benefits including empowering blind people to independently learn about their health and accelerating data entry for large-scale monitoring (e.g., for pandemics such as Covid) by using only a single photograph per LFT test. Accordingly, we explore the abilities of modern foundation vision language models (VLMs) in interpreting such tests.  To enable this analysis, we first create a new labeled dataset with hierarchical segmentations of each LFT test and its nested test result window.  We call this dataset LFT-Grounding. Next, we benchmark eight modern VLMs in zero-shot settings for analyzing these images. We demonstrate that current VLMs frequently fail to correctly identify the type of LFT test, interpret the test results, locate the nested result window of the LFT tests, and recognize LFT tests when they partially obfuscated. To facilitate community-wide progress towards automated LFT reading, we publicly release our dataset at \url{https://iamstuti.github.io/lft_grounding_foundation_models/}
\end{abstract}

%%%%%%%%% BODY TEXT
\section{Introduction}
\label{sec:intro}
Vision-language models (VLMs) have demonstrated impressive zero-shot capabilities in describing images. This development begs a question as to how far such models' abilities extend. We explore VLMs' abilities for a critical medical problem: analyzing Lateral Flow Tests (LFTs)~\cite{budd2023lateral}.  An LFT~\cite{budd2023lateral} is a cost-effective diagnostic tool for rapidly identifying health conditions. 

Our work contributes to the growing interest in automating LFT analysis~\cite{wong2022machine, park2021design, haisma2019head}. The motivation for automated interpretation of LFT results are numerous.  For instance, such a solution could empower blind people to independently learn about their health~\cite{oswal2023study,dcard2022,smith2022, 10226590}, thereby broadening accessibility and inclusivity in healthcare diagnostics.  Automated readers could also accelerate data entry for large-scale monitoring (e.g., for pandemics such as COVID-19) by only requiring the capture of a single photograph per LFT test~\cite{budd2023lateral}. 

\begin{figure}[t]
  \centering
    \includegraphics[width=\columnwidth]{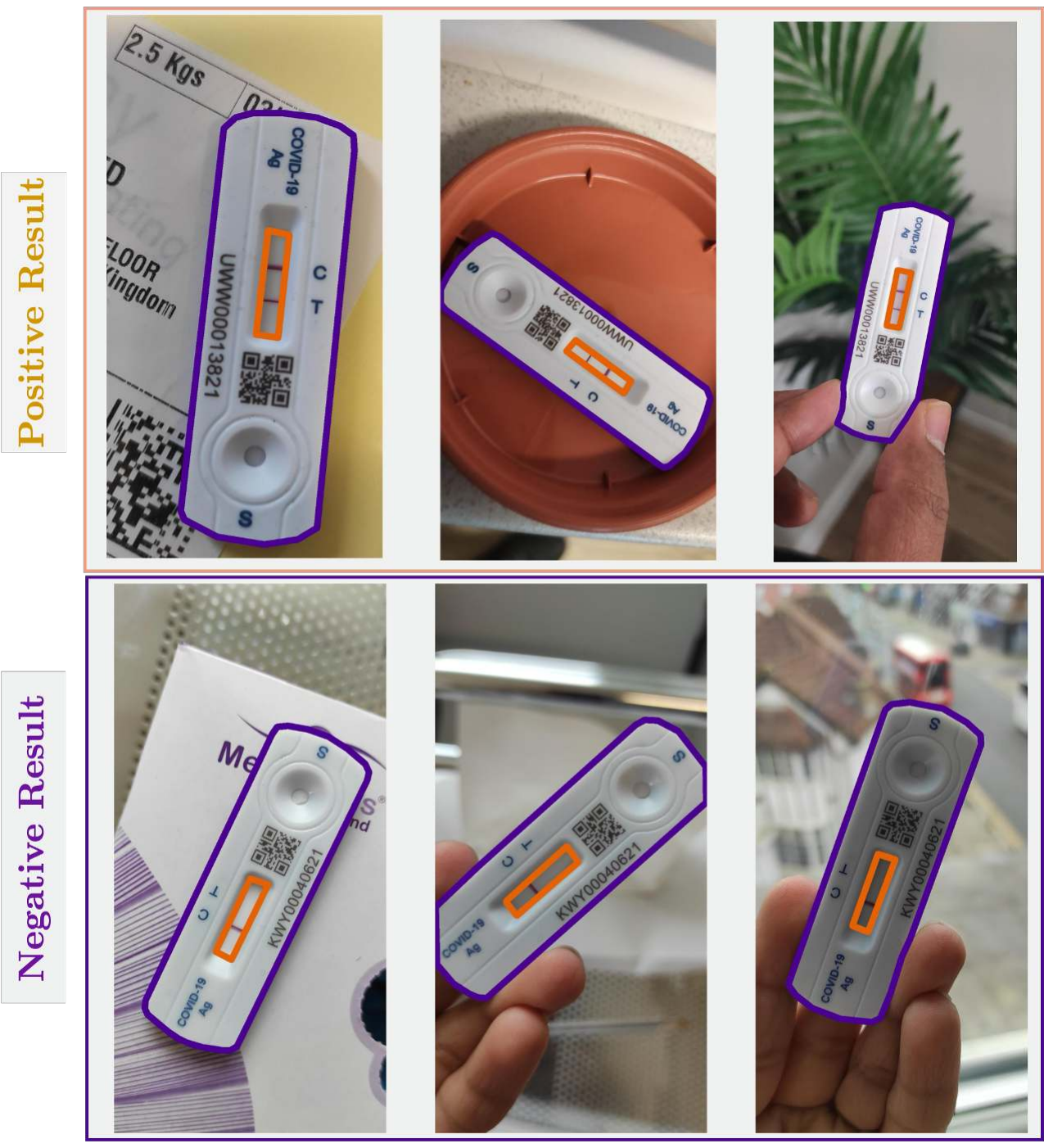}
    \label{fig:short-b}
  \vspace{-4.5em}
  \caption{Examples from our dataset of images showing COVID-19 LFTs with positive results \textit{(first row)} and negative results \textit{(second row)}. We introduce segmentations of each LFT test (indicated in purple) and its test result window (indicated in orange).}
  \label{fig:dataset-example}
\end{figure}

Our first contribution is an LFT-based dataset to enable evaluating models’ predictive performance.  Extending prior work~\cite{turbe2021deep, banathy2021machine, mahdi2023}, we introduce the first dataset that locates the visual evidence used to arrive at a test result interpretation.  Specifically, for a collection of images showing Covid LFTs~\cite{mahdi2023}, we augment each image's label indicating whether the result is positive or negative with segmentations of the test and its nested test result window.  We call the resulting dataset \textbf{LFT-Grounding} to it provides \textbf{groundings} that localize LFT tests and their nested test result windows.  Examples of annotated images are shown in Figure \ref{fig:dataset-example}. 

Our next contribution is to benchmark eight modern VLMs that generate image captions in zero-shot settings to see how effectively they recognize the image contents and reason to those descriptions based on the correct visual evidence.  We find that existing VLMs often struggle to recognize the Covid LFTs, interpret their results, locate the correct visual evidence needed to interpret the test results, and detect partially obscured LFT tests.

We publicly-share our dataset to facilitate future progress on this challenging problem.  Success can benefit other related applications, including automated analysis of other LFT test results including for pregnancy, HIV, and malaria.  Our work also contributes to designing more interpretable solutions, by enabling assessment of the extent to which models reason based on the appropriate visual evidence.
\section{Related Works}
\label{sec:related works}

\paragraph{Lateral Flow Tests (LFTs).}
LFTs have been widely adopted for decentralized testing due to their simplicity, cost-effectiveness, and quick results.  They are used for detecting a broad range of health conditions including pregnancy, Covid, and infectious diseases such as malaria and HIV. Each lateral flow test works by quantifying the presence of a target substance in a liquid sample (e.g., urine, saliva) to determine the presence of a medical condition. Each test achieves this by transporting the sample through pads where nanoparticles with specialized receptors react with the target substance that would indicate a positive result.  The result is a visual signal indicating the strength of that target substance's presence.  Our work will complement the existing work on automatically interpreting LFT results~\cite{wong2022machine, park2021design, haisma2019head}, by offering a new labeled dataset to support richer evaluation alongside the first published analysis of foundation models' zero-shot performance in interpreting LFT test results. 

\paragraph{LFT Datasets.}
A limited number of large-scale, publicly available LFT image datasets exist~\cite{turbe2021deep, banathy2021machine}. Our work is the first to enrich such images with annotations indicating the test result as well as the location of the test and its nested results window. This new dataset can empower researchers to improve the precision of automated visual inspections for LFT images by enabling them to verify models look at the correct visual evidence when making its predictions regarding the test results.

\paragraph{Vision Language Models.}
Many large VLMs ~\cite{li2023blip, dai2023instructblip,li2023monkey, wang2023cogvlm, hanoona2023GLaMM, zhu2023minigpt, cai2023vipllava} have made significant strides in various vision and cross-modality downstream tasks. For example, several models~\cite{li2023monkey, wang2023cogvlm} have shown strong performance on conventional public image captioning datasets like COCO-Captions~\cite{karpathy2015deep}, TextCaps ~\cite{sidorov2020textcaps}, NoCaps ~\cite{agrawal2019nocaps}, and Flickr30k ~\cite{plummer2015flickr30k}, respectively. Such models often achieve their impressive reasoning and generalization capabilities by aligning visual features extracted from images with the input embedding space of power large language models (LLMs) like ChatGPT ~\cite{ChatGPT}, GPT-4 ~\cite{achiam2023gpt}, Vicuna ~\cite{chiang2023vicuna}, and LLaMA ~\cite{touvron2023llama}.  A well-known challenge from such models is the tendency to hallucinate, ignoring the image and instead specifying text the LLM knows should often appear together from training. We rigorously explore modern VLMs to interpret LFT images using specific prompts designed to direct them to locate the LFT tests and the test result window within the images, we aimed to assess the models' ability to draw interpretations from visual cues. Our findings shed light on the current capabilities of VLMs in accurate LFT image analysis, highlighting potential areas of improvement, as it directly impacts the reliability and effectiveness of accurate and unbiased healthcare image interpretation.
\section{LFT-Grounding Dataset}
\label{sec:datasets}
We now introduce our extended version of LFT image dataset~\cite{mahdi2023}, that we call LFT-Grounding to reflect the dataset provide \textbf{g}roundings that localize LFT tests and their nested test result windows.

\cvprsubsection{Dataset Creation}

\paragraph{Source.} 
We extend an existing dataset of 325 Covid Lateral Flow Test (LFT) images acquired in real-world settings, licensed by MIT on Kaggle~\cite{mahdi2023}. Our dataset is restricted to these images due to obstacles in acquiring additional publicly-available images of LFTs.\footnote{Other known images are HIV LFT images in ~\cite{turbe2021deep} and pregnancy LFT images from Adobe Stock~\cite{Adobe}.  However, we are still waiting on our application on February 15,2024 for the former dataset to be approved and the latter restricts their redistribution in the license.} Images are categorized according to their ground-truth labels, distinguishing between positive and negative results for LFT tests. All images contain exactly one clearly-visible, valid LFT test with observable test lines in its result window. 

\paragraph{Annotation Task Design.} 
We created an annotation interface for segmenting parts of LFT images.  It starts with detailed instructions at the top that encompass navigating the interface and completing annotations for images with both positive and negative results, including an annotated example of each scenario.  Then, annotators are walked through annotating five LFT images. For each image, users are first asked, ``Is the Covid test positive? Please indicate ``Yes" or ``No" for the image you are viewing." After collecting the result of the Covid Test, annotators are tasked with outline the entire Covid LFT test within the image. Upon successful demarcation, annotators are then directed to segment the result window. Annotators are first asked, "Can you locate the Covid Test Result on the Covid Test Object?" and, if yes, are then instructed to delineate the result window. The interface enables segmentation by gathering a sequence of clicked points to form a contiguous polygon.  

\paragraph{Annotation Collection.} 
We hired crowdworkers from Amazon Mechanical Turk (AMT) to annotate the data.  We deployed a total of 65 HITs to annotate all 325 images.  We engaged 16 highly skilled annotators from AMT for this task, who already had contributed to previous efforts from our team in the preceding 30 months for large-scale data segmentation.  The authors visually reviewed all submissions to verify they all are high-quality.

\cvprsubsection{Dataset Analysis}
We now characterize our LFT-Grounding dataset with respect to it's overall composition as well as spatial statistics characterizing the Covid tests and their result windows.

\paragraph{Overall Dataset Composition.} 
In Table~\ref{tab: lftdd}, we report for LFT-Grounding the total number of images, number of images showing a positive test result, number of images showing a negative test result, and total number of collected segmentations (recall, each image contains two segmentations with one for the test and one for the result window). Overall, we observe a large dataset imbalance with only 8\% of images indicating a negative result. We suspect this could be attributed in part to people preferring to post images of positive Covid test results online in order to either warn their community or garner emotional support. 

\paragraph{Covid LFT Test Statistics.} We next characterize the segmentations for the test and result window with respect to four metrics:

\begin{itemize}
    \item \textbf{Image Coverage:} fraction of image pixels covered by a segmentation.
    \item \textbf{Test Coverage:} fraction of test pixels covered by the result window.
    \item \textbf{Boundary Complexity (BC):} ratio of a segmentations area to the length of it's perimeter (i.e., the isoperimetric quotient). Values are in $[0,1]$ with 0 representing a highly jagged boundary and 1 representing a perfect circle.
    \item \textbf{Normalized Aspect Ratio (NAR):} Ratio of the shortest side of the segmentation to it's longest side. Values are in $(0,1]$, where values approaching 0 represent a thin segmentation, while 1 represents a perfect square.
\end{itemize}

\begin{figure}[!t]
    \centering
    \includegraphics[width=\columnwidth]{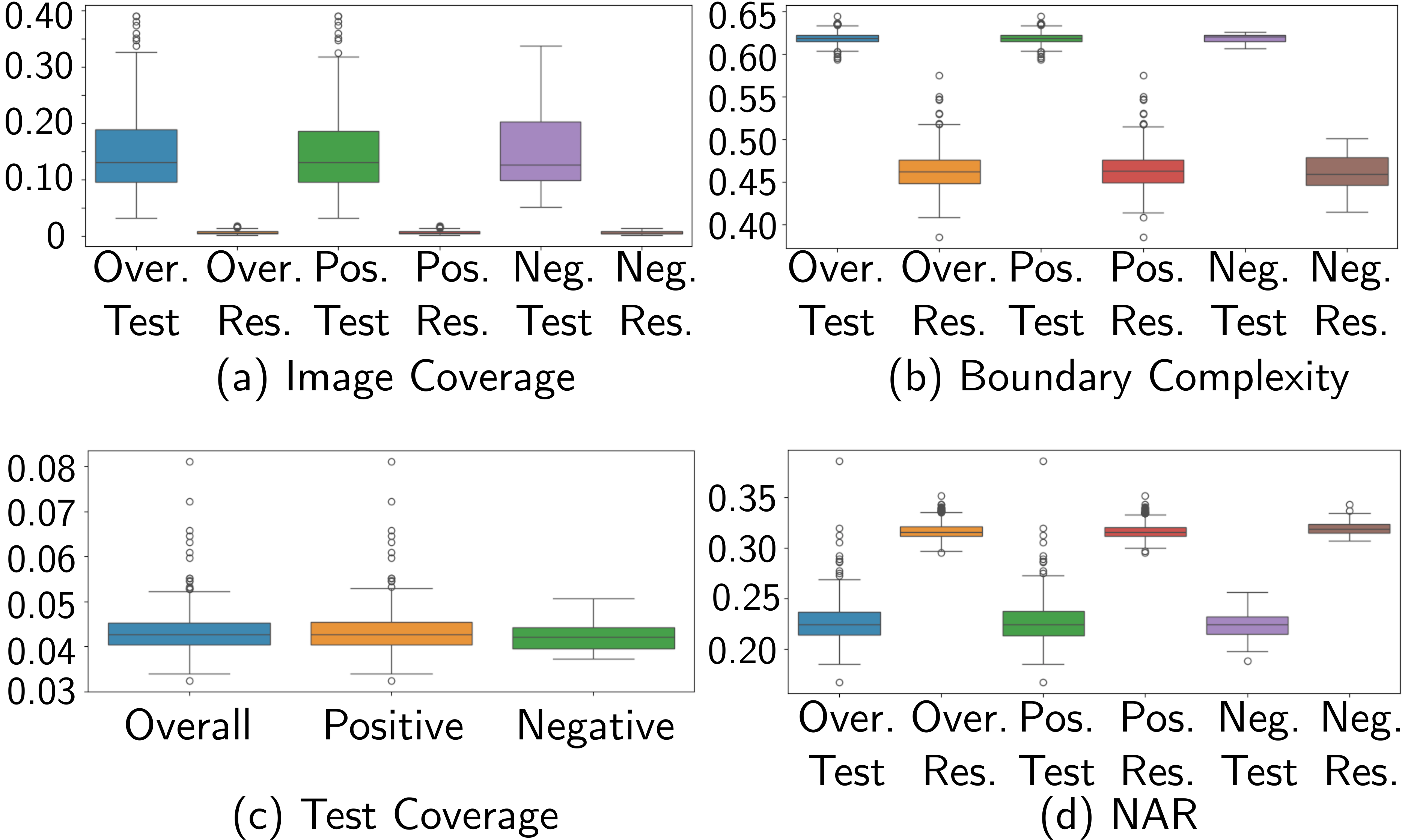}
    \caption{Boxplots for each of the four metrics used to analyze LFT-Grounding. Each boxplot shows the overall results, alongside fine-grained result for positive and negative tests. (a) image coverage; (b) boundary complexity; (c) test coverage; (d) NAR. The lines in each boxplot represent medians, the bottoms and tops of each boxplot represent the 25th and 75th percentiles respectively, whiskers represent most extreme data not considered outliers, and circles represent outliers. (Res.=Result window; Over.=Overall; Pos.=Positive; Neg.=Negative; NAR=Normalized aspect ratio)}
    \label{fig: data_anal}
\end{figure}

\noindent
For all four metrics, we visualize the distribution of scores in Figure~\ref{fig: data_anal}. 

\begin{table}[t!]
\centering
\begin{tabular}{cccc}
    \toprule
     \textbf{\# Images} & \textbf{\# Pos.} & \textbf{\# Neg.}  & \textbf{\# Seg.}\\ \midrule
     325 & 300 & 25 & 650\\ \bottomrule
\end{tabular}
\caption{Composition of LFT-Grounding. From left to right: number of images, number of images showing a positive Covid result, number of images showing a negative Covid result, and total number of segmentations.}
\label{tab: lftdd}
\end{table}

Overall, results for positive and negative results look similar across all metrics.  The one exception is for outliers in positive tests, which we suspect is due to sample imbalance between positive and negative images in our dataset. 

We also observe that the result window occupies a small proportion of pixels in the images as well as within the Covid tests. This highlights a challenge for modern models~\cite{tong2022deep}, as locating small regions is known to be a challenging problem.  Improving automated analysis of such small result windows would be especially valuable for visually impaired users, as the design of result windows is an obstacle for many low-vision users who have some lingering sight yet cannot independently discern information in such small physical regions. 

When observing boundary complexity, we observe the the test itself is more circular than the result window (i.e., has a higher BC score). Intuitively, Covid tests typically have rounded corners, which contributes to having a simpler shape, whereas test windows are typically rectangular with sharp edges, leading to a more jagged shape, as demonstrated in Figure~\ref{fig:dataset-example}. Regions with simple boundaries (e.g., rectangles) may be easier for models to segment due to clear boundaries and less variability in form, compared to more complex shapes (e.g., humans). Finally, we observe that both the result window and rectangular shapes, with Covid tests having a NAR range in $[0.29, 0.35]$ (mean 0.32) and result windows having a range in $[0.17, 0.38]$ (mean $0.23$), indicating that result windows are on average more narrow than the test itself, matching qualitative samples from our LFT-Grounding dataset in Figure~\ref{fig:dataset-example}. As a consequence, models may have trouble accurately localizing result windows as thin objects have shown to be challenging for modern models~\cite{liew2021deep, ke2024segment}.
\section{Algorithm Benchmarking}
\label{sec:algorithm-benchmarking}
We next analyze the performance of modern VLMs at analyzing images in our LFT-Grounding dataset.

%\paragraph{Limitations of Conventional Metrics.} 
%Conventional image captioning metrics such as BLEU, METEOR, ROUGE, CIDEr and SPICE may inadequately capture the semantic similarity between candidate and reference captions,thus,leading to a poor correlation with human judgment, especially for zero-shot models as noted by various studies~\cite{zeng2020intrinsic,cui2018learning, liu2017improved}. \footnote{Conventional image captioning metrics like BLEU, METEOR, ROUGE, and CIDEr primarily measure word overlap between candidate and reference captions, potentially overlooking semantic similarity. SPICE often assigns high scores to lengthy or repetitive sentences, leading to a discrepancy with human evaluation}

\paragraph{Models.}
We chose to benchmark the following eight off-the shelf vision language models in zero-shot settings: BLIP2~\cite{li2023blip}, InstructBLIP~\cite{dai2023instructblip}, MiniGPT-4~\cite{zhu2023minigpt}, CogVLM~\cite{wang2023cogvlm}, Monkey~\cite{li2023monkey}, GLaMM~\cite{hanoona2023GLaMM}, ViP-LLaVA~\cite{cai2023vipllava} and GPT-4V~\cite{achiam2023gpt}.\footnote{While the commercial application, Be My AI~\cite{Paris}, has potential for reading LFT's results, it currently only exists as a mobile application and so is unsuitable for our large-scale evaluation.}  For InstructBLIP and MiniGPT-4, we test both available backbone options.  We summarize these models in Table~\ref{tab:model_visual_input}. 

\begin{table}[b!]
\centering
\resizebox{0.475\textwidth}{!}{%
\begin{tabular}{lccc}
\toprule
% \textbf{Model}        & \begin{tabular}[c]{@{}c@{}}\textbf{Visual}\\\textbf{Input Format}\end{tabular} & \begin{tabular}[c]{@{}c@{}}\textbf{Generate }\\\textbf{Groundings}\end{tabular}  \\ 
\textbf{Model} & \textbf{Visual Input} & \begin{tabular}[c]{@{}l@{}}\textbf{Grounding}\\\textbf{Output}\end{tabular} & \begin{tabular}[c]{@{}l@{}}\textbf{Grounding }\\\textbf{Type}\\\end{tabular}  \\ 
\midrule
BLIP2~\cite{li2023blip} &VP & \textcolor{orange}{$\times$} & -- \\
\hdashline
InstructBLIP(Vicuna)~\cite{dai2023instructblip} &VP & \textcolor{orange}{$\times$} & --\\
InstructBLIP(FlanT5)~\cite{dai2023instructblip} &VP& \textcolor{orange}{$\times$}  & --\\
\hdashline
MiniGPT-4(Vicuna)~\cite{zhu2023minigpt}&VP & \textcolor{orange}{$\times$}  & --\\
MiniGPT-4(Llama)~\cite{zhu2023minigpt}&VP & \textcolor{orange}{$\times$}  & --\\
\hdashline
GPT-4V~\cite{achiam2023gpt}&VP & \textcolor{orange}{$\times$}  & --\\
CogVLM~\cite{wang2023cogvlm}&VP,Coor & \textcolor{purple}{\checkmark}  & \textit{Bbox}\\
Monkey~\cite{li2023monkey}&VP & \textcolor{orange}{$\times$}  & --\\
GLaMM~\cite{hanoona2023GLaMM}&VP,Coor& \textcolor{purple}{\checkmark} & \textit{Mask} \\
ViP-Llava~\cite{cai2023vipllava}& VP,ROI& \textcolor{orange}{$\times$}  & --\\
\bottomrule
\end{tabular}}
\caption{Comparison of different methods in terms of input visual prompt formats and ability to generate groundings as visual output. Input formats include a text prompt along with an image (\textbf{VP}), text prompt specifying image coordinates along with an image (\textbf{Coor}), text prompt along with a region of interest indicated by visual cues such as arrows, bounding boxes, circles, scribbles overlaid onto the input image (\textbf{ROI}). Grounding output formats are bounding box (\textbf{Bbox}) and binary segmentation mask (\textbf{Mask}).}
\label{tab:model_visual_input}
\end{table}

\paragraph{Evaluation Metrics.} 
We use two evaluation metrics to assess a model's abilities to recognize the image contents and to do so based on the correct visual evidence. 

For assessing recognition, we measure \textbf{accuracy} based on each model's ability to both identify that the test is for Covid and provide the correct test result.  We establish a single score per image through two sequential steps of string matching. First, we search for ``Covid-19 Test" (case-insensitive) in each model's generated caption.  If that string is detected, then we perform string matching for \textit{``positive result"} and \textit{``negative result"} to find matches to the ground truth labels.\footnote{Preliminary findings showed better evaluation using ``result" in the string rather than simply searching for ``positive" and ``negative".}  We penalize models as predicting wrong when generated captions include ``positive or negative".\footnote{Preliminary findings gave similar outcomes when instead searching for two strings: ``positive" and ``negative".}

For models that can generate visual groundings (i.e., CogVLM and GLaMM) (Table \ref{tab:model_visual_input}), we evaluate their grounding capabilities by measuring the \textbf{Intersection over Union (IoU)} between the prediction and ground-truth. We do this separately for each entity type of interest, specifically the covid test and its test result window. Due to the different grounding outputs from these models (Table \ref{tab:model_visual_input}), we compare bounding boxes for CogVLM and segmentation masks for GLaMM.

For both measures, we compute the scores across all images. Then, we present results as percentages, where higher values range from 0 to 100 with higher scores signifying better performance. 

%For each salient object, specifically the Covid Test and Test Result Window, we computed the average IoU score across all images in the LFT-Grounding dataset. We report the results as percentages (i.e., IoU $\times$ 100), with higher values indicating better performance.

\paragraph{Test Result Recognition with General Prompts. } 
We first test three prompt formats: $P_{G1}$: ``\textit{Describe the image in detail}”, $P_{G2}$: ``\textit{Describe in detail every object and their parts}”, and $P_{G3}$: ``\textit{Describe the hierarchical parts of the LFT test in the image}". Results are shown in Table \ref{tab:accuracy scores}. 

While most models perform poorly at recognizing the test result, we observe strong performance from GPT-4V with scores ranging from 64\% to 74\% across the three prompts. We suspect GPT-4V's strong performance is due to its more extensive training data, although its proprietary design limits further analysis. 

When comparing the performance of different prompts, we observe marginally improved accuracy scores for the prompt \textit{$P_{G1}$} compared to prompts \textit{$P_{G2}$} and \textit{$P_{G3}$}.  We suspect this is due to the prompt's closer resemblance to those used during the training of these models; e.g., \textit{``Describe this image in detail"} was used when fine-tuning MiniGPT-4~\cite{zhu2023minigpt} and \textit{``Generate the detailed caption in English:"} was used when instruction tuning Monkey~\cite{li2023monkey}). 

\begin{table*}[t]
\centering
\begin{tabular*}{\textwidth}{@{\extracolsep{\fill}}lccccc} 
\toprule
\multicolumn{1}{c}{\multirow{2}{*}{\textbf{Model}}} & \multicolumn{3}{c}{\textbf{General Prompt}} & \multicolumn{2}{c}{\textbf{Specific Prompt}}\\
\multicolumn{1}{c}{} & $\mathbf{P_{G1}}$ & $\mathbf{P_{G2}}$ & $\mathbf{P_{G3}}$ & $\mathbf{P_{S1}}$ & $\mathbf{P_{S2}}$ \\ 
\midrule
BLIP2~\cite{li2023blip} & 0 & 0 & 0 & 0 & 0 \\
\hdashline
InstructBLIP(Vicuna)~\cite{dai2023instructblip} & 0 & 0 & 0 & 0.62 & 0.92 \\
InstructBLIP(FlanT5)~\cite{dai2023instructblip} & 0 & 0 & 0 & 0 & 0 \\
\hdashline
MiniGPT-4(Vicuna)~\cite{zhu2023minigpt} & 0.31 & 0.31 & 1.23 & 4.92 & 2.77 \\
MiniGPT-4(Llama)~\cite{zhu2023minigpt} & 0.31 & 0 & 0 & 3.08 & 2.77  \\
\hdashline
GPT-4V~\cite{achiam2023gpt} & 74.15 & 64.62 & 67.38 & 62.24 & 74.77  \\
Monkey~\cite{li2023monkey} & 2.77 &  0.62 & 0 & 0.62 & 0 \\
CogVLM~\cite{wang2023cogvlm}& 23.69 & 15.08 & 4.62  & 30.77 & 7.38 \\
GLaMM~\cite{hanoona2023GLaMM} & 0 & 0 & 0 & 0 & 0 \\
ViP-Llava~\cite{cai2023vipllava} & 3.38 & 2.77 & 0 & 23.69 & 8.62 \\
\bottomrule
\end{tabular*}
\caption{Test result recognition performance from models prompted in a zero-shot setting. $P_{G/S}$ refers to different general and specific prompts respectively, as described in (Section \ref{sec:algorithm-benchmarking}). }
\label{tab:accuracy scores}
\end{table*}

\paragraph{Test Result Recognition With Prompts Specifying the Test Type.} 
We next prompt the model with additional information about the test type as follows: ``\textit{$P_{S1}$: Describe in detail every part of Covid test in the image}", ``\textit{$P_{S2}$: Describe the hierarchical parts of Covid test in the image}". Results are reported in the Table \ref{tab:accuracy scores}.  We observe an average improvement of $\sim$5\% in accuracy scores when we notify models about the test type as opposed to general prompts. We suspect specifying the test type helps models know that common language patterns are to specify a test result when reporting the presence of a Covid test.

%\paragraph{Test Result Recognition With Prompts Indicating What to Look.}
%We next assess the models' ability in interpreting Covid LFT images in our dataset if they are notified explicitly where to look in the given image. To achieve this, we first provide the ground-truth test location through the prompt: \textit{“$P_{L2}$: Describe if there is a COVID test in the image, if so, what is the result?”}. We report the results in the Table \ref{tab:accuracy scores}. We observe prompting with the appropriate visual evidence in the images yields an improved outcome.

%Next, we evaluate the improvement in models' ability to comprehend the Covid LFT image when it is also provided with additional information about the ground-truth label (i.e., positive or negative) as a description for the Covid Test object. We utilize the prompt: \textit{``$P_{A1}$:Can you provide a description of \(<expr>\) COVID-19 test in the image?”,} we replace \textit{\(<expr>\)} with the ground-truth test label (i.e., positive or negative). 

\begin{figure}[t]
    \centering
    \includegraphics[width=0.95\columnwidth]{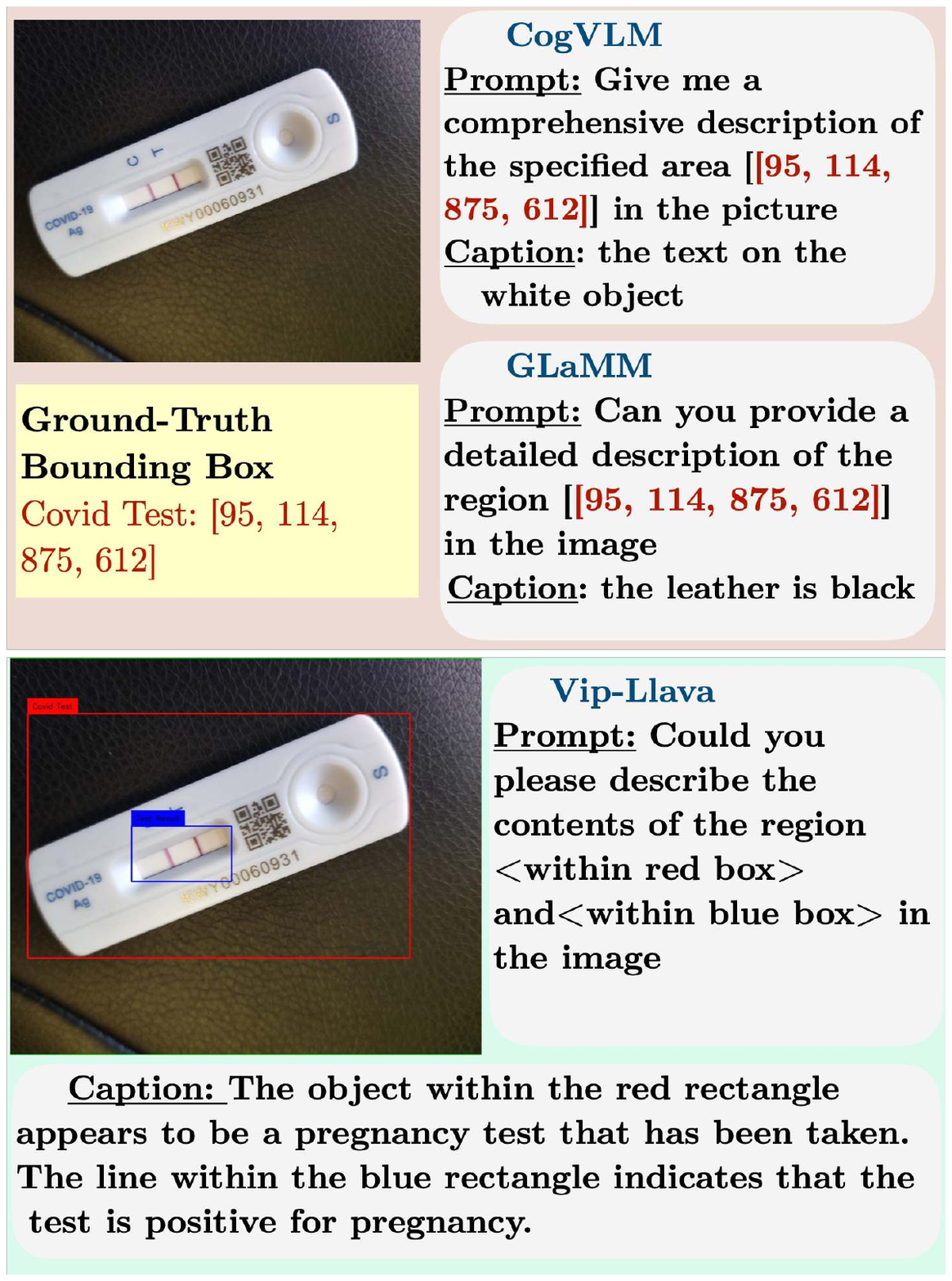}
    \vspace{-0.3em}
    \caption{Example of generated captions by the models when they are notified in the prompts of the Covid test's location.}
    \label{fig:bbox prompt p1}
\end{figure}

\paragraph{Test Result Recognition When Notifying Model's About the Covid Test's Location.}
We next assess the models' perfomance when they are notified where to look in the given image. We feed bounding box coordinates of the Covid test's location (derived from its ground-truth segmentation) to the models that accept coordinates as inputs to explore the upper bound of what these models can achieve. We use predefined prompts for visual inputs from the original documentations of CogVLM~\cite{Thudm}, GLaMM~\cite{hanoona2023GLaMM}, and VipLlava~\cite{cai2023vipllava}.\footnote{For CogVLM, we use the prompt template: ``\textit{Give me a comprehensive description of the specified area [[x0,y0,x1,y1]] in the picture}". For GLaMM, we use: ``\textit{Can you provide a detailed description of the region [[x0,y0,x1,y1]] in the image}". Unlike CogVLM and GLaMM, which accept bounding boxes through text prompts, Vip-Llava processes the image with overlaid multiple bounding boxes, accompanied by a text prompt specifying the visual cues (e.g., `` \textless red bounding box \textgreater", ``green circle"). The prompt used for Vip-Llava is: \textit{``Could you please describe the contents of the region \textless within red box \textgreater and \textless within blue box \textgreater in the image"} (where \textless within red box \textgreater refers to the Covid LFT Test object's bounding box and \textless within blue box \textgreater refers to the LFT Test Result Window).} Quantitative results are reported in Table \ref{tab:accuracy per ground-truth bbox aux prompt} and a qualitative result is shown in Figure \ref{fig:bbox prompt p1}. 

We find that all models generally perform worse on our dataset when given ground-truth bounding box coordinates through prompts. While Vip-Llava misinterprets the Covid LFT image as `pregnancy test' in 99\% of the cases, as exemplified in the Figure \ref{fig:bbox prompt p1}, it  slightly surpasses its counterparts (i.e., CogVLM and GLaMM). We believe Vip-Llava's improved performance stems from its inherent model design~\cite{cai2023vipllava}. Unlike CogVLM and GLaMM, which are limited to understanding information conveyed through textual prompts, Vip-Llava is uniquely designed to interpret both the overlaid visual markers (specifically, bounding boxes that identify Covid Test and Test Result Window in our case) on the image and the accompanying prompt. This dual-layered approach enables Vip-Llava to generate better descriptions for the LFT images.

\begin{table}[t]
\centering
\begin{tabular}{@{\extracolsep{\fill}}l|l}
\toprule
\textbf{Model} & \textbf{Bbox Prompt} \\ 
\midrule
CogVLM~\cite{wang2023cogvlm}& 0\\
GLaMM~\cite{hanoona2023GLaMM}& 0  \\
ViP-Llava~\cite{cai2023vipllava}&0.31 \\
\bottomrule
\end{tabular}
 \caption{Accuracy metrics for correctly indicating positive and negative test results in generated captions with respect to Covid LFT images in the dataset, when models operate in their zero-shot setting using auxiliary prompts which provide ground-truth bounding box coordinates for the Covid test.}
  \label{tab:accuracy per ground-truth bbox aux prompt}
\end{table}

\begin{table}[t]
\centering
\begin{tabular}{lll} 
\toprule
\textbf{Model} & \textbf{IoU$_{LFT-Test}$} & \textbf{IoU$_{Result-Window}$}  \\ 
\midrule
CogVLM~\cite{wang2023cogvlm}    & 28   & 16.09  \\
GLaMM~\cite{hanoona2023GLaMM}  & 97.59& 4.69    \\
\bottomrule
\end{tabular}
\caption{Overall performance of CogVLM~\cite{wang2023cogvlm} and GLaMM~\cite{hanoona2023GLaMM} in locating the Covid LFT test as well as its nested result window for all images in our dataset (mean value reported).}
\label{tab: Avg IOU}
\end{table}

\paragraph{Evaluating VLMs with Grounding Abilities in Zero-shot Setting for LFT Recognition}
We assess the visual grounding proficiency of two models supporting this capability: CogVLM and GLaMM. For CogVLM, which supports bounding box predictions, we design the prompt: \textit{``Please describe the \textless entity \textgreater in detail and provide its coordinates [[x0, y0, x1, y1]]”}. For GLaMM, which supports segmentation predictions, we design the prompt: \textit{``Can you please segment \textless entity \textgreater in the given image”}. We use for \textless entity \textgreater both ``Covid Test" and ``Covid Test Result Window". Results are shown in Table \ref{tab: Avg IOU}.\footnote{We exclude performance scores for recognizing the test result because in most instances the models didn't output a string description of the image.}

\begin{figure}[t]
    \centering
    \includegraphics[width=\columnwidth]{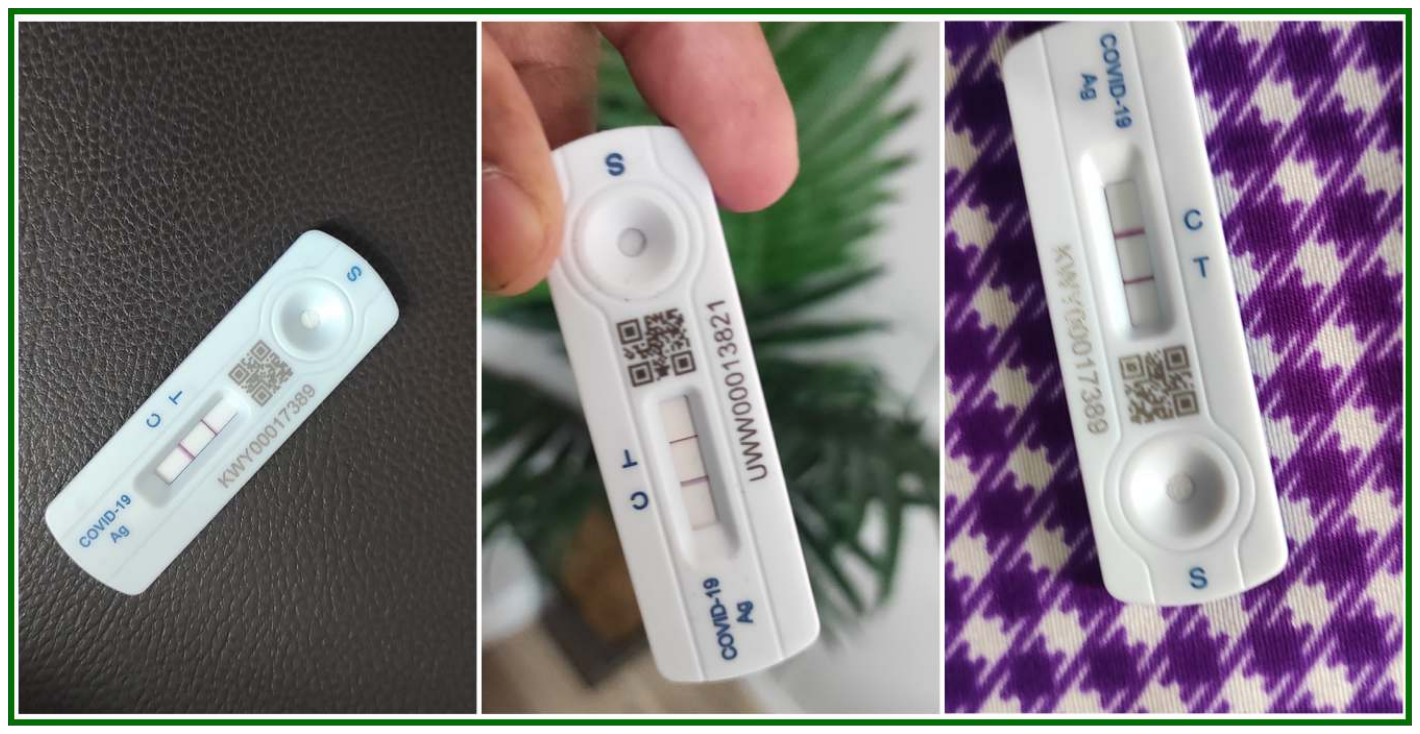}
    \vspace{-6.5em}
    \caption{Examples when CogVLM did not generate bounding box predictions for both the Covid test and its result window.}
    \label{fig:cogvlm bbox failure}
\end{figure}

For locating the Covid tests, we observe higher IoU scores from GLaMM than CogVLM. We suspect superior performance of GLaMM stems in part from providing fine-grained pixel-level object groundings rather than coarse bounding boxes like CogVLM.  Additionally, we attribute this to the observation that in ~7\% of cases when CogVLM is prompted for object ``Covid Test", and in $< 2\%$ of cases when prompted for ``Test Result Display Window", it fails to provide the grounding coordinates for the specified salient object in the prompt. A commonality shared in these cases is either the LFT test is partially obscured (e.g., from being held in a hand) or the LFT test is placed on a dark or highly textured background, as exemplified in Figure \ref{fig:cogvlm bbox failure}. 

When observing performance gaps between average IoU scores for the predicted bounding box for the Covid test and test result window, we see that both models struggle to identify the nested test result window within the LFT images. Specifically, the performance decrease is $\sim$42.50\% for CogVLM and $\sim$95.31\% for GLaMM. We suspect this decline in performance is due to VLMs' limited grasp in recognizing hierarchical decomposition within objects as well as in interpreting small and thin entities. This finding underscores a valuable future direction in improving the visual grounding capabilities at multiple decomposition levels that include smaller entities.

\begin{figure*}[t]
    \centering
    \hspace*{-1.1cm}    
    \includegraphics[width=\textwidth]{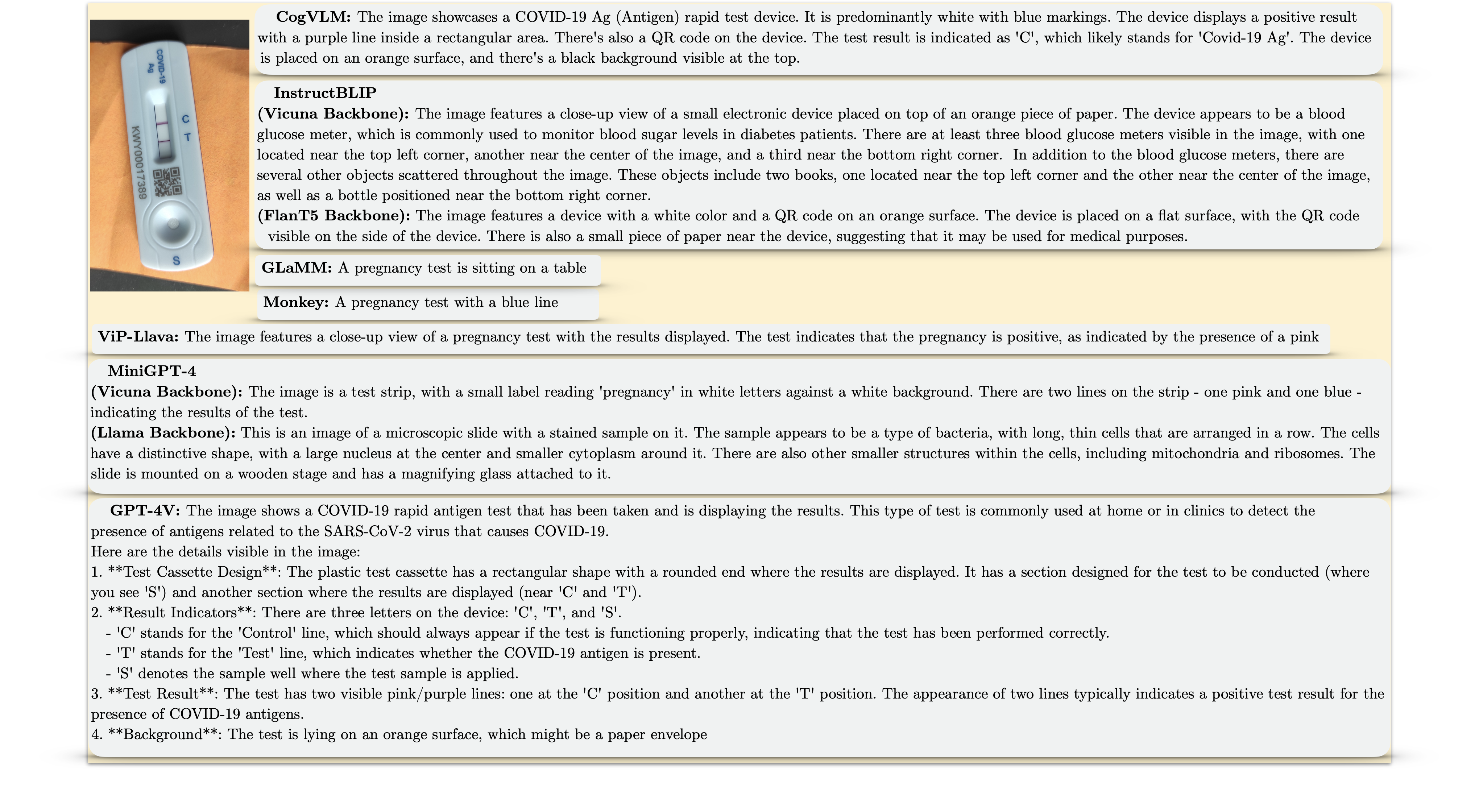}
    \vspace{-2.0em}
    \caption{Models' predictions using the overall top-performing prompt, \textit{``Describe the image in detail"}, for a test with a positive result.}
    \label{fig:gen prompt p1}
\end{figure*}

\paragraph{Qualitative Results}
We display the captions generated by the models for the top-performing prompt \textit{``Describe the image in detail"} in Figure \ref{fig:gen prompt p1}. 

\begin{figure*}[t]
    \centering
    \includegraphics[width=0.9\textwidth]{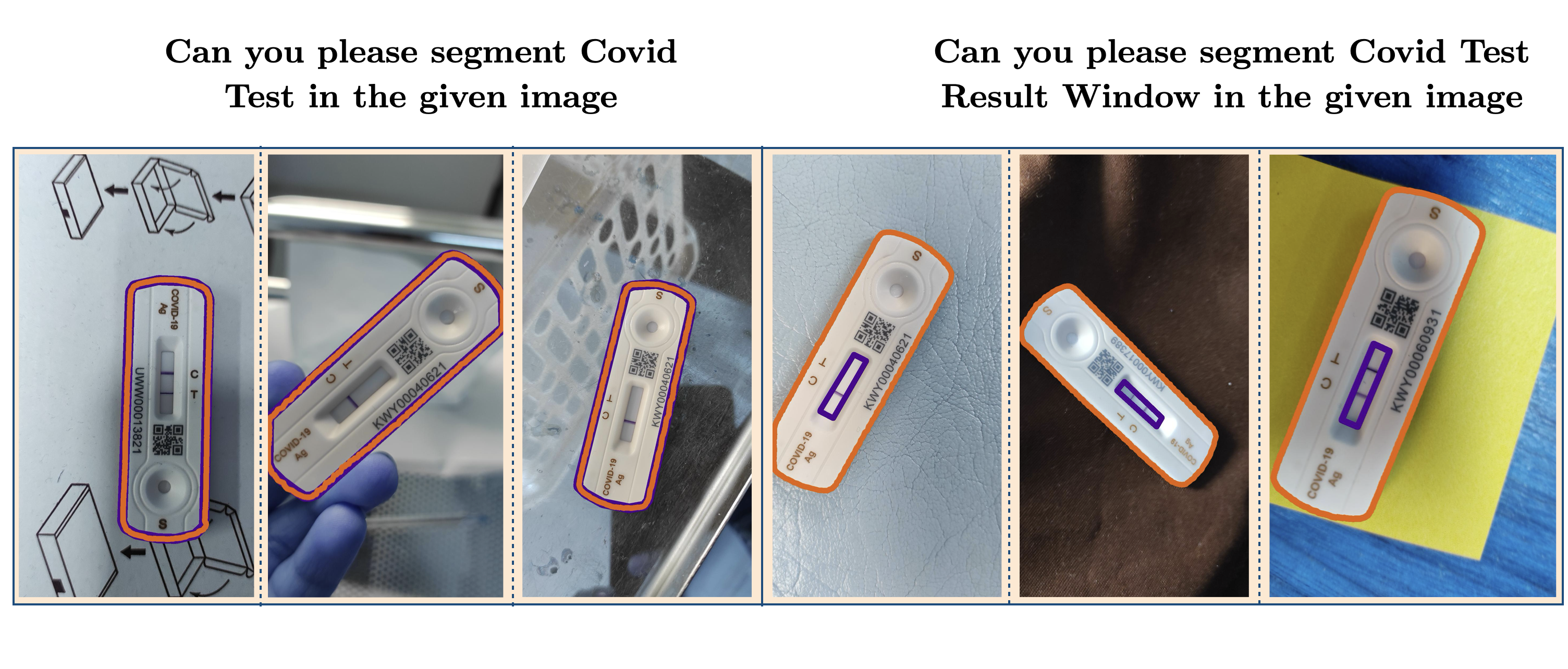}
    \vspace{-2.0em}
    \caption{Examples of ground-truth (purple overlay) and GLaMM predictions (orange overlay) for locating the Covid test as well as its nested result window.}
    \label{fig:seg glamm results}
\end{figure*}

Upon visual inspection, we observe a common mistake from models is misidentifying the Covid LFT image as a ``pregnancy test," ``thermometer," or ``blood glucose monitor." While these errors may stem from model hallucinations, we also suspect this poor performance is due to training datasets lacking sufficient diversity of LFT test images. 

We also show an example of the predictions for the Covid test and the nested test result window by GLaMM in Figure \ref{fig:seg glamm results}.  This reinforces our quantitative finding that performance declines between from the Covid test to the locating its nested result window.

\section{Conclusions}
\label{sec:conclusions}
We introduce LFT-Grounding to catalyze research into improving the zero-shot generalization capabilities of state-of-the-art models for automatic interpretation of LFT images. We also benchmark eight state-of-the-art models in zero-shot setting to highlight their current status for this challenging problem.  Our work underscores opportunities for future work, including on resolving how to acquire LFT images for a wider range of health conditions as well as on improving the performance of automatic models in interpreting results and achieving this by conveying the visual evidence it used to arrive at that interpretation.  An important step for future work will include expanding our test result categorizations to also support recognizing ``invalid" test results. We publicly share our dataset to spur community effort to facilitate future extensions of this work.

% This gives rise to the need for LFT dataset containing images corresponding to a wide range of infections and conditions

% Mention as future work handling detection of "invalid" test results.

\vspace{-1em}
\paragraph{\bf Acknowledgments.}
This project was supported in part by a National Science Foundation SaTC award (\#2148080) and gift funding from Microsoft AI4A. Josh Myers-Dean is supported by a NSF GRFP fellowship (\#1917573). We thank Bo Xie, Tom Yeh, and Dan Larremore for helping inspire this research direction.

\newpage
%\clearpage

%%%%%%%%% REFERENCES
{\small
\bibliographystyle{ieee_fullname}
\bibliography{egbib}
}

%-------------------------------------------------------------------------
% Use \appendix command to start the appendix section
% \clearpage
% \appendix
% \input{sec/06-Appendix}

\end{document}